\documentclass[letterpaper]{article}

\usepackage{natbib,alifeconf, graphicx,amsmath,amssymb, caption, fancyhdr}  
\usepackage{url,hyperref,cleveref}
\usepackage{booktabs}
\title{Adaptive Exploration in Lenia with Intrinsic Multi-Objective Ranking}

\author{
    Niko Lorantos$^{1}$ \and
    Lee Spector$^{1,2}$\\
    \mbox{}\\
    $^1$University of Massachusetts Amherst, USA \\
    $^2$Amherst College, USA\\
    nlorantos@umass.edu
} 

\begin{document}

\maketitle

\begin{abstract}
Artificial life aims to understand the fundamental principles of biological life by creating computational models that exhibit life-like properties. Although artificial life systems show promise for simulating biological evolution, achieving open-endedness remains a central challenge. This work investigates mechanisms to promote exploration and unbounded innovation within evolving populations of Lenia continuous cellular automata by evaluating individuals against each other with respect to distinctiveness, population sparsity, and homeostatic regulation. Multi-objective ranking of these intrinsic fitness objectives encourages the perpetual selection of novel and explorative individuals in sparse regions of the descriptor space without restricting the scope of emergent behaviors. We present experiments demonstrating the effectiveness of our multi-objective approach and emphasize that intrinsic evolution allows diverse expressions of artificial life to emerge. We argue that adaptive exploration improves evolutionary dynamics and serves as an important step toward achieving open-ended evolution in artificial systems.
\end{abstract}



\section{Introduction}
The emergence of complex lifeforms in nature can be attributed to billions of years of evolutionary innovation. This process relies on the ability to continually produce novelty without limit, a characteristic known as open-endedness (OE). However, achieving OE in artificial life (ALife) systems remains a central challenge that is fundamental to replicating biological complexity and discovering new forms of intelligence. Most artificial systems fail to enable the degree of unbounded innovation seen in biology while even the definition of OE remains debated, highlighting the need for concrete definitions and robust metrics to appropriately identify open-ended behavior in ALife \citep{Stepney2021, Banzhaf2016,Stepney2024}. 

Despite these definitional challenges, this work serves as a practical step toward achieving OE in ALife. We investigate how multi-objective optimization of intrinsic fitness objectives serves as a method for enabling adaptive exploration in evolutionary simulations. We rank individuals against each other to encourage innovation without restricting the scope of emergent behaviors. Through multi-objective ranking, we enable thorough exploration of individuals that represent different tradeoffs in the search space. We find that this intrinsic multi-objective approach produces a diverse population of artificial lifeforms and may enable sustained innovation and open-ended evolution in ALife.
\subsection{Lenia and Leniabreeder}
Lenia \citep{Chan2018}, a family of continuous cellular automata, is a promising platform for exploring OE in ALife. Lenia generalizes Conway's Game of Life by using continuous states and differentiable update rules based on convolution. It models complex, life-like patterns and emergent behaviors through iterative applications of these rules on a grid. Lenia patterns have demonstrated capabilities such as self-organization, homeostatic regulation, locomotion, entropy reduction, growth, adaptability, and evolvability \citep{Chan2020}. This positions Lenia as the ideal testbed for investigating the fundamental mechanisms that drive complexity and novelty in evolutionary systems.

The Leniabreeder framework \citep{Faldor2024} further enhances Lenia by integrating Quality-Diversity (QD) algorithms to facilitate the selection of diverse and high-performing patterns in evolving populations of Lenia. Leniabreeder utilizes a Variational Autoencoder (VAE) to represent patterns in a latent descriptor space, enabling the comparison of phenotypes. The AURORA QD algorithm \citep{Grillotti2022} extends this approach by utilizing diversity in the VAE latent space to guide selection. Leniabreeder’s unsupervised learning techniques and automated feature discovery are fundamental to enabling OE, allowing the system to adapt without human intervention. Our work utilizes the open-source Leniabreeder framework and AURORA QD algorithm to evolve populations of Lenia patterns.

\subsection{Open-Endedness}
\cite{Banzhaf2016} have proposed a framework in which OE can be defined as three types of novelty:

\begin{tabular}{|l|l|}
\hline
\textbf{Type} & \textbf{Definition} \\
\hline
Type-0 & Variation within the model. \\
Type-1 & Innovation that changes the model. \\
Type-2 & Emergent changes to the meta-model. \\
\hline
\end{tabular}

In this paper, we define OE as the continual production of novelty of these types \citep{Banzhaf2016,Stepney2021,Taylor2016}. Our approach specifically aims to promote Type-0 variation and Type-1 innovation, utilizing Leniabreeder’s VAE latent space to capture behavioral diversity within the population. However, OE remains challenging to precisely define and measure as true open-ended behavior will effectively move outside any predefined measures of OE, making attempts to quantify OE potentially trivial \citep{Stepney2024}. We will instead focus on exploring underlying mechanisms that promote the continual production of novelty described to exhibit open-ended behavior by the above definition.
\subsection{Intrinsic Evolution}
Systems that utilize internal mechanisms to guide evolution are considered to be intrinsic, as the fitness landscape is shaped by the system's current state rather than progressing toward a predetermined goal. Intrinsic evolution enables complexity and diversity to emerge, mirroring the open-ended dynamics of natural evolution. This approach is promising in its ability to guide systems toward meaningful innovation, and has been shown to promote OE in foundational ALife literature.

Novelty search \citep{Lehman2011}, for instance, abandons objectives and instead rewards behavioral novelty. This approach demonstrates efficacy in tasks such as maze navigation and biped locomotion. By abandoning explicit fitness objectives, novelty search enables a continual increase in complexity. This concept is extended in Flow Lenia, a mass-conservative extension of Lenia, where \cite{Plantec2023} enable multi-species simulations to drive intrinsic evolutionary processes through competition and symbiosis. Furthermore, \cite{Reinke2019} explore intrinsically motivated goal exploration processes (IMGEPs) in self-organizing systems like Lenia, emphasizing the automated discovery of diverse patterns. Their IMGEP-OGL algorithm utilizes deep auto-encoders and achieves efficiency comparable to pretrained systems, highlighting the power of unsupervised learning in uncovering novel self-organized structures. 

Together, these approaches emphasize the potential for evolution driven by intrinsic processes and autonomous exploration to replicate the open-ended creativity of natural systems. Decoupling evolution from fixed objectives has proven to aid the emergence of life-like behaviors in artificial environments. We utilize this philosophy in the design of our multi-objective fitness mechanism, enabling divergent evolution guided by exploration.

\begin{figure*}[htbp]
    \includegraphics[width=\textwidth]{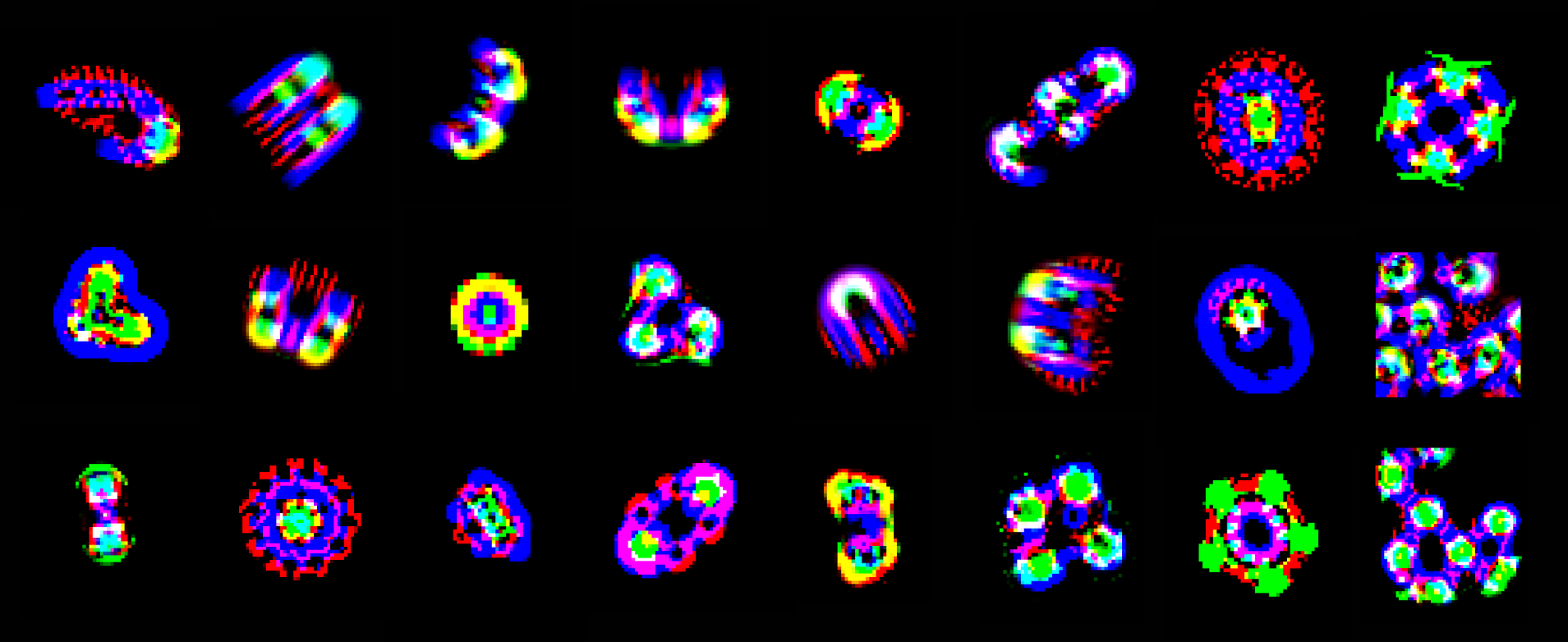}
  \caption{Patterns evolved in Leniabreeder using intrinsic multi-objective ranking.}
  \label{fig:teaser}
\end{figure*}

\section{Methodology}
We evaluate individuals with respect to three objectives: \textit{homeostasis}, \textit{distinctiveness}, and \textit{population sparsity}. These objectives capture notions of behavioral diversity within the population and promote individuals with complex internal structure in underexplored regions of the descriptor space. Their intrinsic nature allows the system to evolve in accordance with its own evolutionary activity. 

\textit{Homeostasis} rewards stable artificial lifeforms, comparable to those in biological life. We calculate latent variance of an individual, measuring stability in the latent space representation across \textit{n} timesteps
\begin{equation}
f_1 = -\frac{1}{n}\sum_{i=1}^{n}|\vec{z}_i - \bar{\vec{z}}|
\label{eq:homeostasis}
\end{equation}
where $\vec{z_i}$ are latent encodings and $\bar{\vec{z}}$ is their mean. \textit{Distinctiveness} encourages novelty relative to preexisting individuals. We calculate latent mean distance, measuring divergence from average behavior
\begin{equation}
f_2 = |\bar{\vec{z}}-\mathbb{E}[\bar{\vec{z}}]|
\label{eq:distinctiveness}
\end{equation}
where $\bar{\vec{z}}$ is the mean latent encoding of an individual. \textit{Population sparsity} rewards individuals in underexplored regions of the descriptor space. We calculate the density of the descriptor space using a radial basis function (RBF) kernel
\begin{equation}
f_3 = -\sum_{a \in \mathrm{archive}} \exp \left( -\frac{|\vec{d} - \vec{d_a}|^2}{2\sigma^2} \right)
\label{eq:sparsity}
\end{equation}
where $\vec{d}$ is the descriptor vector and $\sigma$ is the kernel width.

Together these intrinsic objectives formulate the multi-objective problem; balancing them is crucial for producing meaningful populations of Lenia patterns through evolution. Optimizing for homeostatic regulation alone offers no adaptive or explorative benefits, while optimizing solely for exploration would result in patterns that explode or dissipate, eliminating the biological relevance of our study. However, these objectives together show promise for enabling greater exploration of self-regulating patterns in the search space. The clear tradeoffs between these objectives highlight the potential for multi-objective optimization to loosely guide evolution toward the discovery of innovative artificial lifeforms.

We implement a domination count fitness mechanism to rank members of the population against those in the archive --- the set of previously generated patterns --- based on these objectives. Domination count ranking is simple yet elegant in its ability to promote adaptive exploration, as ranking individuals against each other exhibits the relative selection needed to continually produce novelty. Domination count of an individual \textit{x} is calculated as
\begin{equation}
    d(x)=|\{y \in A \mid y \prec x\}|
    \label{eq:domination}
\end{equation}
where \textit{A} is the set of archived individuals. An individual \textit{y} is said to be dominated by \textit{x}, denoted $y \prec x$, if \textit{x} is better than or equal to \textit{y} in all objectives, and strictly better in at least one objective. The final fitness score of an individual is their negative domination count, so solutions dominated by fewer archive members are considered fitter. This approach maintains a diverse set of individuals representing different trade-offs between objectives, thereby fostering greater exploration of the solution space. By ranking individuals against each other, there is no pressure on the system to converge toward a final goal, but rather exploration of the search space is promoted \citep{Lehman2011}. 

We compare our multi-objective ranking approach to the single-objective homeostasis fitness function in \Cref{eq:homeostasis}. This comparison enables investigation into the effects of explorative objectives and domination count ranking. We conducted 50 trials for each approach, evolving 2500 generations with batch sizes of 256 and a repertoire size of 1024. We recorded the mass, repertoire variance, and complexity of individuals throughout the evolutionary process. Mass is calculated as the zeroth spatial moment, or total sum of all cell values across all channels, representing the total life content in a pattern. Repertoire variance is calculated as the latent variance across solutions for the last timestep, giving a measure of phenotypic diversity. Finally, we calculate complexity as the gzip compression size of phenotypes across all frames. This serves as an approximation of Kolmogorov complexity, the size of the shortest program that can reproduce a pattern's behavior.

\section{Results and Discussion}
We report the average mass, repertoire variance, and complexity of individuals across the final populations. Multi-objective ranking displayed increased mass, variance, and compressibility compared to the homeostasis implementation. Table \ref{tab:quant} summarizes the quantitative results from our experiments.
\begin{table}[h]
\centering
\begin{tabular}{|c|c|c|c|}
\hline
\textbf{Metric} & \textbf{Homeostasis} & \textbf{Multi-Objective} & \textbf{$\Delta$} \\
\hline
Mass & 3.268 & \textbf{3.292} & +0.73\% \\
\hline
Variance & 1.093 & \textbf{1.103} & +0.91\% \\
\hline
Complexity & \textbf{3.861} & 3.820 & -1.06\% \\
\hline
\end{tabular}
\caption{Results of homeostasis and multi-objective fitness mechanisms across all individuals in final repertoires.}
\label{tab:quant}
\end{table}

These results indicate enhanced evolutionary dynamics that foster greater exploration of the search space in the multi-objective approach. The deviations in average mass, variance, and complexity between the two approaches each demonstrated statistical significance (all p $<$ 0.001) using pooled two-sample t-tests. The observed increase in average mass suggests that our approach successfully preserved life content and potentially adaptive traits. Similarly, the increase in repertoire variance confirms greater diversity within the populations evolved through intrinsic multi-objective ranking. The observed reduction in complexity, coupled with increased mass and variance, points to the emergence of patterns exhibiting greater modular internal structure and avoidance of trivial complexification as the system continually innovates without converging on a fixed goal. These results highlight the effectiveness of intrinsic multi-objective ranking in enabling adaptive exploration without compromising on meaningful homeostatic regulation. 

\begin{figure}
    \centering
    \includegraphics[width=3.1in]{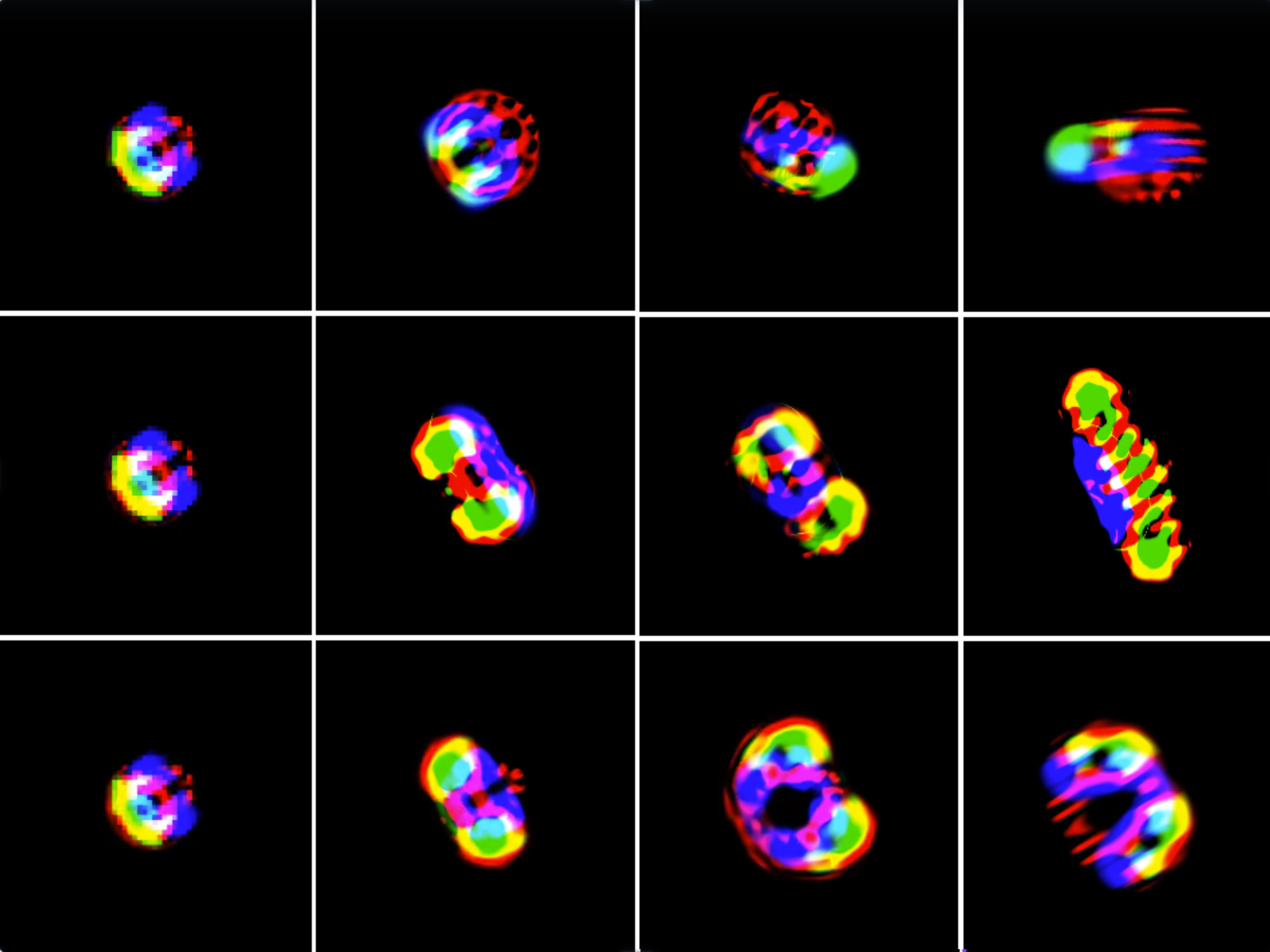}
    \caption{Temporal progression of three patterns starting from the “Aquarium” phenotype (pattern id 5N7KKM) \citep{Chan2020, Faldor2024}.
    }
    \label{fig:frames}
\end{figure}

\begin{figure*}[htbp]
    \includegraphics[width=\textwidth]{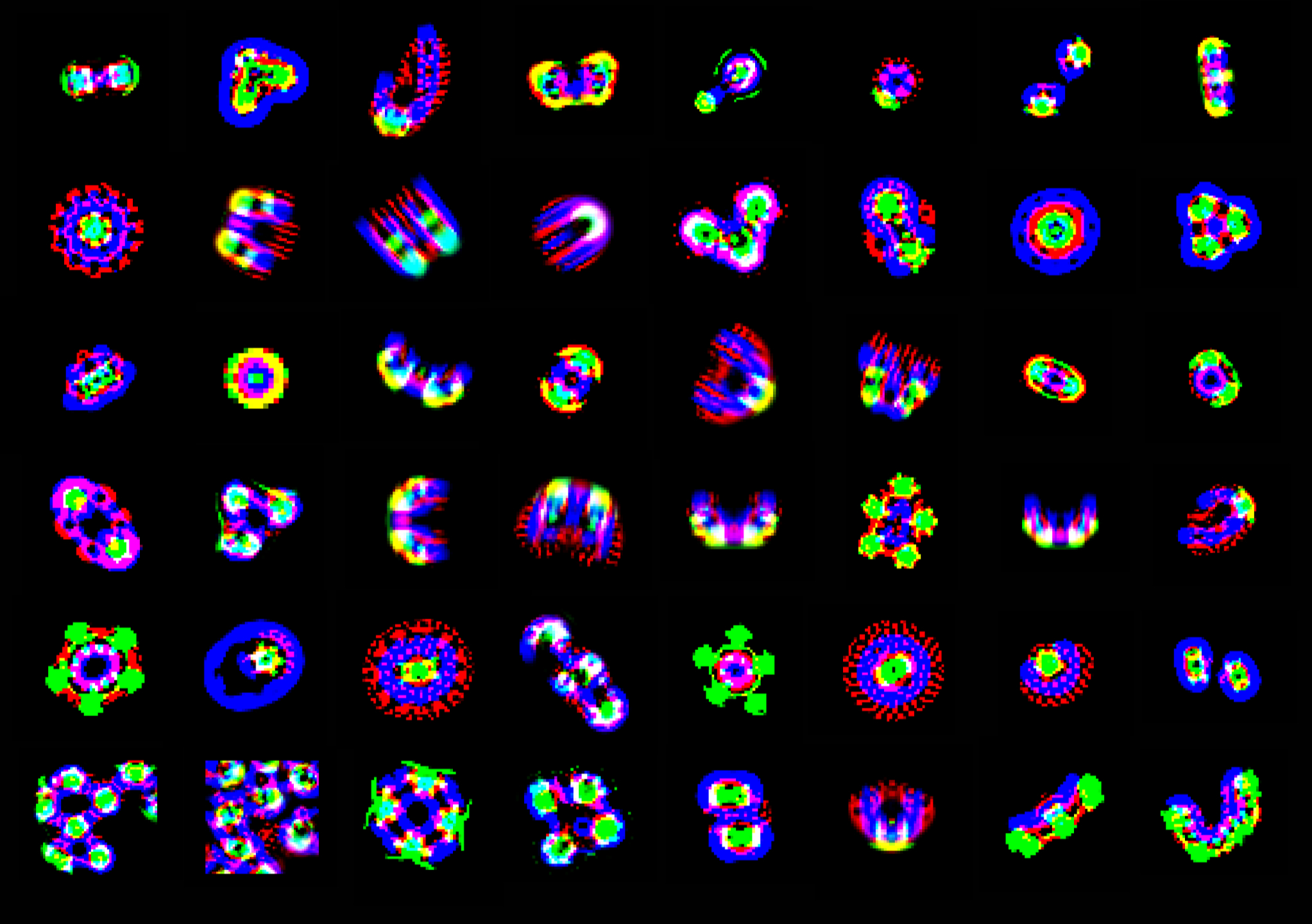}
    \caption{Columns 1-4 display patterns evolved through multi-objective ranking, while columns 5-8 display patterns evolved through single-objective homeostasis fitness and the AURORA QD algorithm.}
    \label{fig:patterns}
\end{figure*}

We observed subtle qualitative differences between individual patterns evolved through both approaches, and attribute diversity in the homeostasis approach to the AURORA QD algorithm. However, multi-objective ranking yielded patterns with greater structural variety and enhanced internal modularity, displayed in Figure \ref{fig:patterns}. Figure \ref{fig:frames} showcases the behavior of such patterns over multiple timesteps. These patterns maintain complex internal structure through time and vary from each other, demonstrating the potential for intrinsic multi-objective ranking to promote greater variation and innovation within evolving ALife populations. By simultaneously optimizing for homeostasis, distinctiveness, and sparsity, the system generates novelty while preserving stability. This balance allows a population of diverse artificial lifeforms to emerge through adaptive exploration.

\section{Future Work}
Our findings highlight several promising directions for future research. We intend to apply established measures of open-ended dynamics, such as Bedau and Packard’s evolutionary activity test \citep{Bedau1997} and the MODES toolbox \citep{Dolson2019}, to quantify OE in our system. We posit that more comprehensive measures of OE will offer important insights into the open-ended nature of various evaluation mechanisms in evolutionary computation. Additionally, we aim to explore more advanced fitness objectives, such as homeodynamic regulation in place of homeostasis, potentially fostering artificial lifeforms with more intricate and adaptive internal structures. We plan to introduce environmental pressures to assess the system’s adaptive complexity, hypothesizing that intrinsic multi-objective ranking will outperform a baseline QD algorithm in discovering innovative solutions. Lastly, given the relatively small scale of our experiments, we anticipate that intrinsic multi-objective ranking at scale will display emergent capabilities analogous to those observed in evolutionary biology.

\section{Conclusion}
This study demonstrates that intrinsic multi-objective ranking enables adaptive exploration in artificial life, serving as a step toward achieving open-endedness in evolutionary computation. We find that intrinsic fitness objectives encourage variation and innovation in evolving populations of Lenia. By ranking individuals against each other, we enable greater diversity and exploration of the search space unrestricted by bounded fitness functions. The theoretical advantages of intrinsic multi-objective ranking in promoting exploration, coupled with experimental results, indicate that our approach can promote the continual production of novelty in evolutionary search.

Our contribution extends beyond performance gains to offer a novel methodological framework for enabling open-endedness by promoting autonomous exploration and decreasing reliance on manual feature design. By aligning fitness objectives with intrinsic characteristics of living systems, we foster ongoing innovation akin to natural evolution. Our approach applies explorative mechanisms inspired by biological evolution to artificial life, bringing us closer to achieving open-endedness in evolutionary computation.

\section{Acknowledgements}
We thank members of the Amherst College PUSH lab for stimulating conversations that helped us develop the ideas described in this paper. This material is based upon work supported by the National Science Foundation under Grant No. 2117377. Any opinions, findings, conclusions, or recommendations expressed in this publication are those of the authors and do not necessarily reflect the views of the National Science Foundation. This work was also performed in part using high-performance computing equipment obtained under a grant from the Collaborative R\&D Fund managed by the Massachusetts Technology Collaborative.

\footnotesize
\bibliographystyle{apalike}
\bibliography{references}

\end{document}